\documentclass[runningheads]{llncs}

\usepackage{eccv}

\usepackage{eccvabbrv}
\usepackage{soul}
\usepackage{graphicx}
\usepackage{booktabs}
\usepackage[export]{adjustbox}
\usepackage{tcolorbox}
\tcbuselibrary{breakable}

\usepackage[accsupp]{axessibility}  %

\usepackage{hyperref}

\usepackage{orcidlink}

\definecolor{Gray}{gray}{0.9}
\definecolor{mygreen}{rgb}{0.0, 0.5, 0.0}
\definecolor{myred}{rgb}{0.8, 0.25, 0.33}
\definecolor{myblue}{rgb}{0.19, 0.55, 0.91}
\definecolor{uclablue}{rgb}{0.15, 0.45, 0.68}
\definecolor{ucladblue}{rgb}{0.0, 0.33, 0.53}
\definecolor{ucladdblue}{rgb}{0.0, 0.23, 0.36}
\definecolor{uclagold}{rgb}{1.0, 0.82, 0.0}
\definecolor{ucladgold}{rgb}{1.0, 0.78, 0.17}
\definecolor{ucladdgold}{rgb}{1.0, 0.72, 0.11}
\definecolor{boxgreen}{rgb}{0.02, 0.66, 0.02}
\definecolor{boxred}{rgb}{0.66, 0.1, 0.1}
\definecolor{boxblue}{rgb}{0.01, 0.01, 0.73}

\usepackage{etoc}
\usepackage{lipsum}
\usepackage{wrapfig}
\usepackage{multicol}
\usepackage{mdframed}

\usepackage{mathtools}

\usepackage[utf8]{inputenc}   %
\usepackage[T1]{fontenc}      %
\usepackage{url}              %
\usepackage{amsfonts}         %
\usepackage{nicefrac}         %
\usepackage{microtype}
\usepackage{graphicx}
\usepackage{amsmath,amssymb,mathbbol}
\usepackage{algorithmic}
\usepackage[linesnumbered,ruled,vlined]{algorithm2e}
\usepackage{acronym}
\usepackage{enumitem}
\usepackage{balance}
\usepackage{xspace}
\usepackage{setspace}
\usepackage[skip=3pt,font=small]{subcaption}
\usepackage[skip=3pt,font=small]{caption}
\usepackage[dvipsnames]{xcolor}
\usepackage{booktabs,tabularx,colortbl,multirow,array,makecell}
\usepackage{pifont}
\usepackage{xr-hyper}
\usepackage{tcolorbox}
\tcbuselibrary{breakable}

\usepackage{siunitx}   %
\usepackage{soul}   %
\usepackage{listings}
\definecolor{codegreen}{rgb}{0,0.6,0}
\definecolor{codegray}{rgb}{0.5,0.5,0.5}
\definecolor{codepurple}{rgb}{0.58,0,0.82}
\definecolor{backcolour}{rgb}{0.95,0.95,0.92}

\lstdefinestyle{mystyle}{
    backgroundcolor=\color{backcolour},   
    commentstyle=\color{codegreen},
    keywordstyle=\color{magenta},
    numberstyle=\tiny\color{codegray},
    stringstyle=\color{codepurple},
    basicstyle=\ttfamily\footnotesize,
    breakatwhitespace=false,         
    breaklines=true,                 
    captionpos=b,                    
    keepspaces=true,                 
    numbers=left,                    
    numbersep=5pt,                  
    showspaces=false,                
    showstringspaces=false,
    showtabs=false,                  
    tabsize=2
}

\makeatletter
\renewcommand{\paragraph}{%
  \@startsection{paragraph}{4}%
  {\z@}{0ex \@plus 0ex \@minus 0ex}{-1em}%
  {\hskip\parindent\normalfont\normalsize\bfseries}%
}
\makeatother

\definecolor{gblue}{HTML}{4285F4}
\definecolor{gred}{HTML}{DB4437}
\definecolor{ggreen}{HTML}{0F9D58}

\definecolor{mygray}{gray}{.92}

\newcommand{\supp}{\textit{Appendix}\xspace}

\acrodef{llm}[LLMs]{large language models}
\acrodef{vla}[VLA]{vision-language-action}
\acrodef{vl}[VL]{vision-language}
\acrodef{ocot}[O-CoT]{Object-centric Chain-of-Thought}
\acrodef{cot}[CoT]{Chain-of-Thought}
\acrodef{lvlm}[LVLM]{large vision-language models}

\newcommand{\method}{\textit{VideoAgent}\xspace}

\renewcommand{\emph}[1]{\textit{#1}}
\newcommand{\toolA}[0]{\includegraphics[scale=0.05,valign=c]{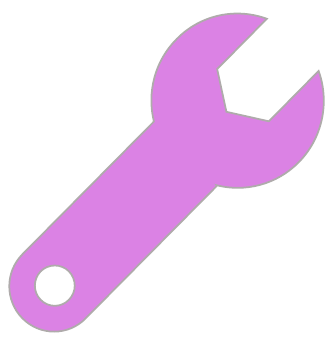}}
\newcommand{\toolB}[0]{\includegraphics[scale=0.05,valign=c]{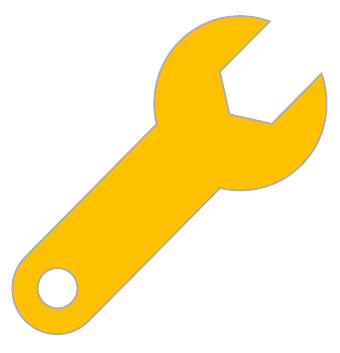}}
\newcommand{\toolC}[0]{\includegraphics[scale=0.05,valign=c]{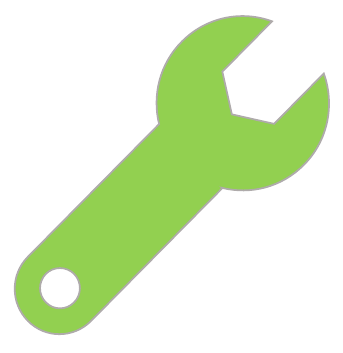}}
\newcommand{\toolD}[0]{\includegraphics[scale=0.05,valign=c]{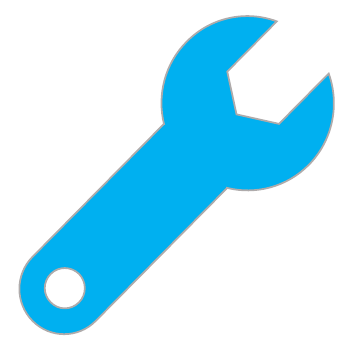}}
\newcommand{\videoagent}[0]{\includegraphics[scale=0.04,valign=c]{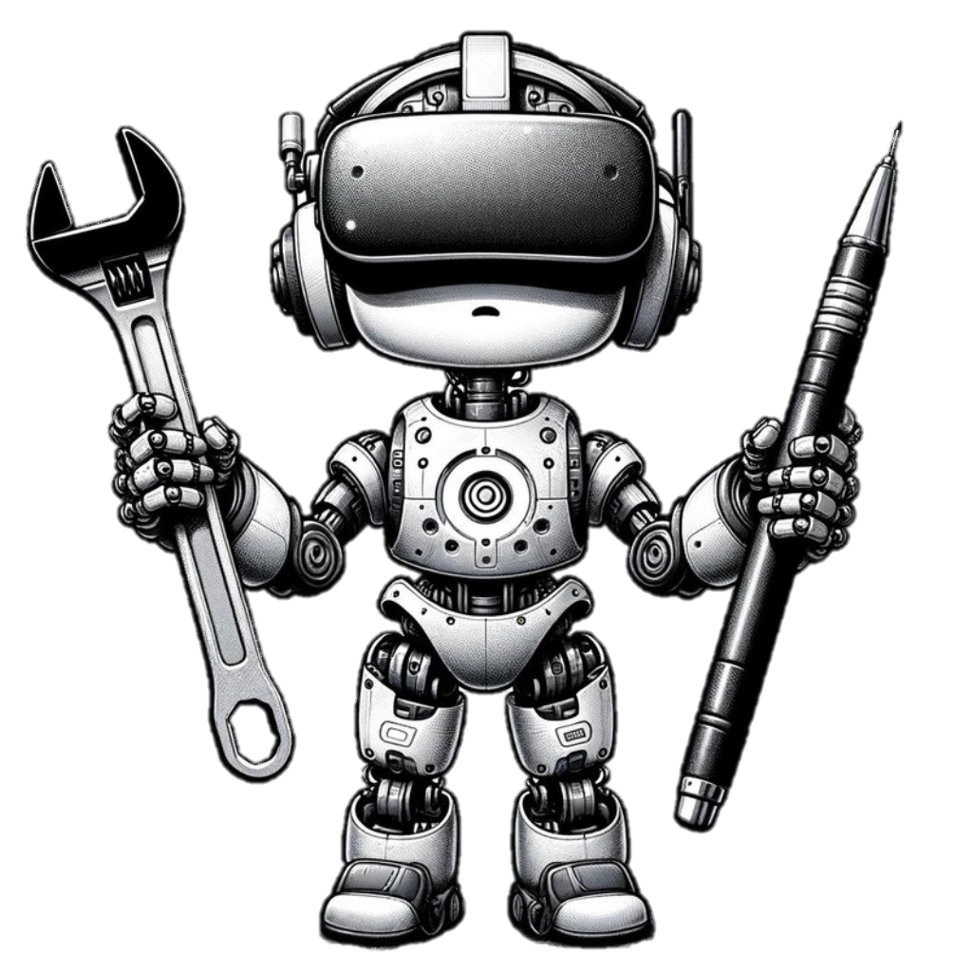}}
\begin{document}

\title{\videoagent \method: A Memory-augmented \\Multimodal Agent for Video Understanding} 

\titlerunning{\videoagent VideoAgent}

\author{Yue Fan$^\star$\inst{1}\orcidlink{0009-0005-5635-4320} \and
Xiaojian Ma$^\star$$^\dag$\inst{1}\orcidlink{0000-0001-5609-3822} \and
Rujie Wu\inst{1,2}\orcidlink{0009-0001-6426-1248} \and
Yuntao Du\inst{1}\orcidlink{0009-0001-4709-4003} \and
Jiaqi Li\inst{1}\orcidlink{0000-0003-1318-5123} \and
Zhi Gao\inst{1,3}\orcidlink{0000-0002-4424-4352} \and
Qing Li$^\dag$\inst{1}\orcidlink{0000-0003-1185-5365}}

\authorrunning{Fan and Ma et al.}

\institute{State Key Laboratory of General Artificial Intelligence, BIGAI, Beijing, China\and
School of Computer Science, Peking University, Beijing, China\and
School of Intelligence Science and Technology, Peking University, Beijing, China\\
\email{\{maxiaojian,liqing\}@bigai.ai}\\
\url{https://videoagent.github.io}
}

\maketitle
\renewcommand{\thefootnote}{}
\footnotetext{$^\star$Equal contribution.}
\footnotetext{$^\dag$Corresponding authors.}
\renewcommand{\thefootnote}{\arabic{footnote}}
\setcounter{footnote}{0}

\begin{abstract}
We explore how reconciling several foundation models (large language models and vision-language models) with a novel unified memory mechanism could tackle the challenging video understanding problem, especially capturing the long-term temporal relations in lengthy videos. In particular, the proposed multimodal agent \method: 1) constructs a structured memory to store both the generic temporal event descriptions and object-centric tracking states of the video; 2) given an input task query, it employs tools including video segment localization and object memory querying along with other visual foundation models to interactively solve the task, utilizing the zero-shot tool-use ability of LLMs. \method demonstrates impressive performances on several long-horizon video understanding benchmarks, an average increase of 6.6\% on NExT-QA and 26.0\% on EgoSchema over baselines, closing the gap between open-sourced models and private counterparts including Gemini 1.5 Pro. The code and demo can be found at \url{https://videoagent.github.io}.

\keywords{video understanding \and LLMs \and tool-use \and multimodal agents}
\end{abstract}

\section{Introduction}\label{sec:intro}

Understanding videos and answering free-form queries (question answering, content retrieval, \etc) remains a major challenge in computer vision and AI~\cite{alayrac2022flamingo,team2023gemini,jxma_vlm_multimodal_2023,wang2024lstp,suris2023vipergpt,liu2023visual,mangalam2024egoschema,song2023moviechat,jia2020lemma,jia2022egotaskqa}. Notably, much of the recent progress has achieved by the end-to-end pretrained large transformer models, especially those are developed upon the powerful \ac{llm}~\cite{wang2024lstp,jxma_llm_vla_vlm_mas_multiagent_2023,song2023moviechat,lin2023video}, \ie multimodal \ac{llm}. However, there have been increasing concerns about their capabilities to handle long-form videos with rich events and complex spatial-temporal dependencies~\cite{tapaswi2016movieqa,miech2019howto100m,wu2021towards,korbar2023text,han2022temporal,jia2022egotaskqa,jia2020lemma}. Specifically, the computation, especially memory cost could grow significantly and even become prohibitively expensive when processing lengthy videos~\cite{wiles2022compressed,team2023gemini}. Also, the self-attention mechanism could sometimes struggle to capture the long-range relations~\cite{tay2020long}. These issues have hindered further advancement in applying sophisticated foundation models to video understanding.

\begin{figure}[t!]
    \centering
    \includegraphics[width=0.9\textwidth]{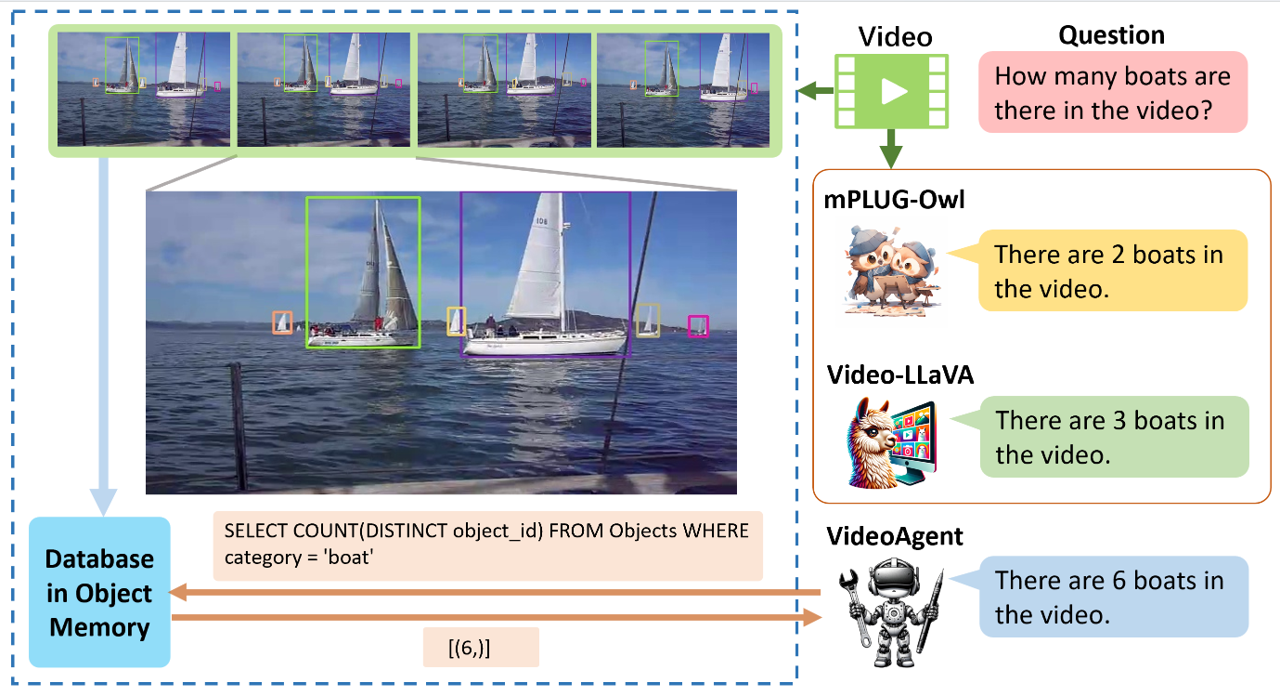}
    \caption{A comparison between \method and end-to-end multimodal LLMs on video question answering. Without a unified memory as a structured representation for videos, end-to-end models could struggle with capturing basic spatial-temporal details, especially when asked about objects and lengthy videos. While \method can utilize a curated set of tools to perform sophisticated queries about the \textit{temporal memory} (not shown) and \textit{object memory}, and respond with the correct answer.}
    \label{fig:boat_example}
\end{figure}

More recently, thanks to the tool-use capabilities of \ac{llm}~\cite{schick2024toolformer,gao2023clova}, there has been rapid development of a new class of multimodal understanding approaches: \textit{multimodal agents}~\cite{gupta2023visual,suris2023vipergpt,liu2023visual,wu2023visual}. The key idea is prompting \ac{llm} into solving the multimodal tasks by invoking several \textbf{tool} foundation models (object detection, visual question answering, \etc) interactively. These methods have great potential as they are mostly training-free and flexible with tool sets. However, extending them to video understanding, especially on long-form videos is \textbf{non-trivial}. Simply adding video foundation models as tools could still suffer from the computation cost and attention limitation issues~\cite{lin2023video,song2023moviechat}. Other research has explored more sophisticated prompting strategies with better tools~\cite{maaz2023video,wang2023lifelongmemory,yang2024doraemongpt}, but they usually lead to complicated pipelines and the performances of these methods still fail to match their end-to-end counterparts possibly due to a lack of video-specific agent design.

In this paper, we introduce a simple yet effective LLM-based multimodal tool-use agent \method for video understanding tasks. Our \textbf{key insight} is to represent the video as a structured unified memory, therefore facilitating strong spatial-temporal reasoning and tool use of the LLM, and matching/outperforming end-to-end models, as shown in \cref{fig:boat_example}. Our memory design is \textbf{motivated} by the principle of being minimal but sufficient: we've found that the overall event context descriptions and temporally consistent details about objects could cover the most frequent queries about videos. As a result, we design two memory components: 1) \textit{temporal memory}, which stores text descriptions of each short (2 seconds) video segment sliced from the complete video; 2) \textit{object memory}, where we track and store the occurrences of objects and persons in the video. To answer a query, the LLM will decompose it into several subtasks and invoke the tool models. The unified memory is centered around by the following tools: \toolA\xspace\textit{caption retrieval}, which will return all the event descriptions between two query time steps; \toolB\xspace\textit{segment localization}, which retrieves a short video segment of a given textual query by comparing it against the event descriptions within the temporal memory; \toolC\xspace\textit{visual question answering}, which answers a question given a retrieved video segment; \toolD\xspace\textit{object memory querying}, which allows sophisticated object state retrieval from the object memory using SQL queries. Finally, the LLM will aggregate the response of the interactive tool use and produce an answer to the input query. 

We conduct extensive evaluations of \method on several video understanding tasks, including free-form query localization with Ego4D NLQ~\cite{grauman2022ego4d}, generic video question answering with WorldQA~\cite{worldqa} and NExT-QA~\cite{xiao2021next}, and egocentric question answering with EgoSchema~\cite{mangalam2024egoschema}, a recent benchmark focusing on complex questions about long-form videos. We compare \method against both the canonical end-to-end multimodal LLMs and other multimodal agents. Results demonstrate the advantages of \method: on averaged increasing 6.6\% on NExT-QA and 26.0\% on EgoSchema over baselines. Our further investigation has examined the role played by the unified memory and tool selection.

To summarize, our contributions are as follows:
\setlength{\leftmargini}{0.85em}
\begin{itemize}[topsep=0pt]
\item We propose a unified memory mechanism to build structured representations for long-form videos, including a \textit{temporal memory} that stores segment-level descriptions and an \textit{object memory} that tracks the state of objects in the video.
\item Based on the unified memory, we design \method, an LLM-powered multimodal agent for video understanding. It decomposes the input task queries and interactively invokes tools to retrieve information from the memory until reaches the final response.
\item We perform thorough evaluations of \method on multiple video understanding benchmarks against both end-to-end multimodal LLMs and multimodal agent baselines, demonstrating the effectiveness of \method. The additional ablation analysis further confirms the crucial design choices we've made.
\end{itemize}

\section{VideoAgent}
\begin{figure*}[t]
    \centering
    \includegraphics[width=0.98\textwidth]{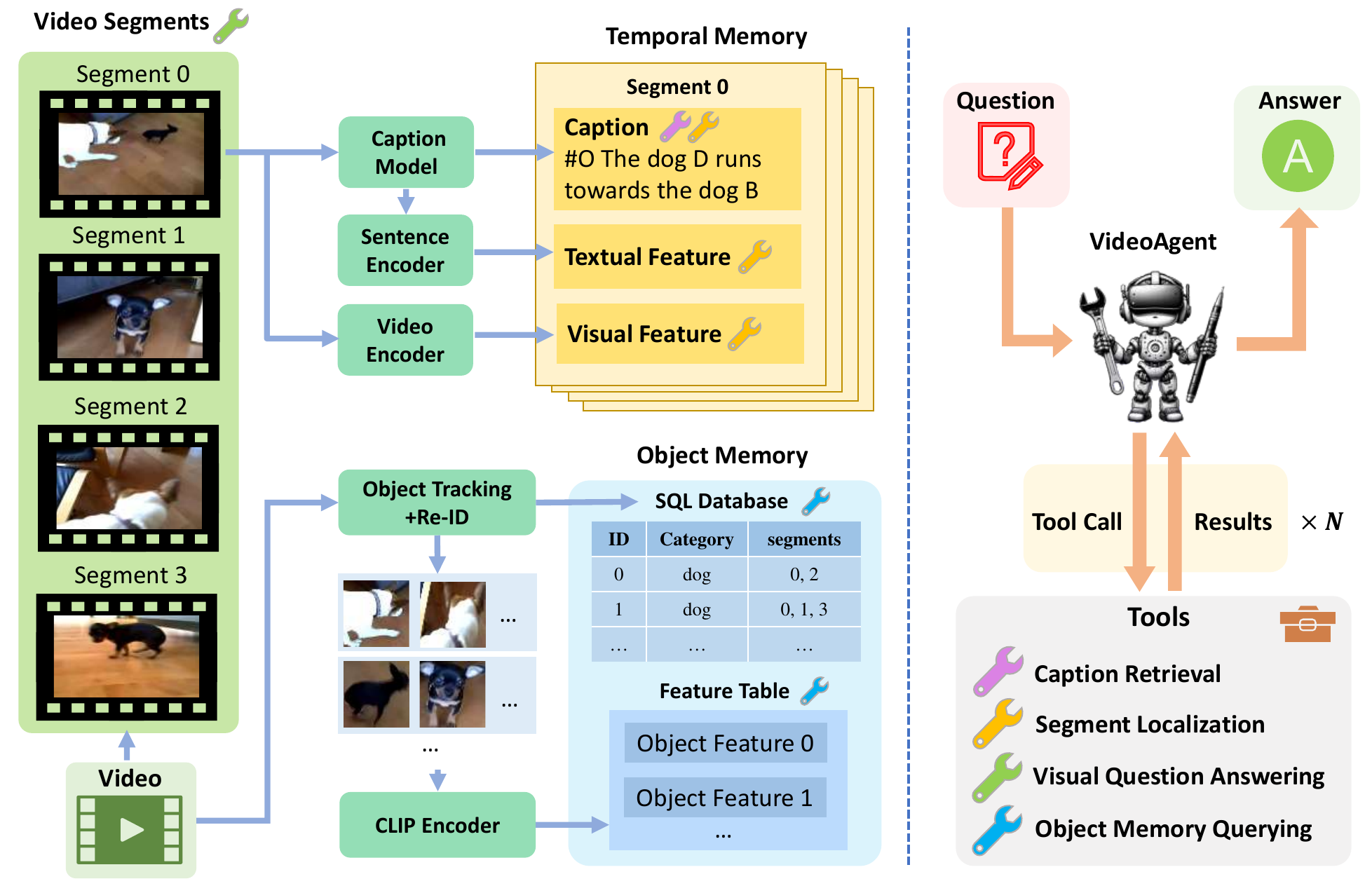}
    \caption{An overview of \method. Left: We first translate an input video into structured representations: a temporal memory and an object memory; Right: the LLM within \method will be prompted to solve the given task by interactively invoking tools (\toolA\toolB\toolC\toolD). Our considered tools primarily work with the memory (\eg \toolA\xspace interacts with the caption part of the temporal memory while \toolD\xspace looks up the object memory).}
    \label{fig:teaser}
\end{figure*}

\subsection{Overview}

We illustrate the proposed \method in \cref{fig:teaser}. It begins with converting the input video into a unified representation: \textit{temporal memory} (\cref{sec:method_t_mem}) and \textit{object memory} (\cref{sec:method_object_mem}). For any incoming task, it interactively invokes tools to collect information from the memory and the raw video segments, and ultimately produces a response (\cref{sec:method_tool}). The memory construction and task-solving (inference) procedures are summarized in \cref{alg:memory} and \cref{alg:inference}, respectively.

\subsection{Temporal Memory $\mathcal{M}_T$}\label{sec:method_t_mem}
The temporal memory is designed to store overall event context descriptions and features of videos. Given $n$ video segments $[v_1, \dots, v_n]$ sliced from a video $V$, we extract video segment caption $s_\text{caption}$, video segment feature $e_\text{video}$ and the caption text embedding $e_\text{caption}$:

\noindent\textbf{Video segment caption.} We use a pretrained video captioning model called LaViLa~\cite{lavila} to produce captions for each video segment. Specifically, it takes $4$ frames from a 2-second segment to produce a short caption sentence. 
Typical LaViLa captions can be "\#C C cuts a wood with a wood cutter" and "\#O The man Y pushes a stroller on the road with his left hand", where "\#C" and "\#O" is used to denote whether the caption sentence is about the camera wearer or someone other than the camera wearer, therefore making LaViLa captions effective in both egocentric and generic videos.

\noindent\textbf{Video segment feature and caption feature.} To obtain the video segment feature, we adopt the video encoder of ViCLIP~\cite{viclip} to encode video segments. We uniformly sampled $10$ frames from each video segment as the input to ViCLIP, and save the generated feature of the segment. For the caption feature, we choose \texttt{text-embedding-3-large}\footnote{\url{https://platform.openai.com/docs/guides/embeddings}} offered by OpenAI to compute the embedding of the video segment caption we obtained from LaViLa.

\subsection{Object Memory $\mathcal{M}_O$}\label{sec:method_object_mem}

In addition to the general video event context stored in the temporal memory, it is also crucial to explicitly capture the temporally consistent details: \eg the presence of people, objects, and the surroundings, \etc. The intuition is that most queries about videos are object(person)-related; therefore, the occurrences of objects (and people) are tracked and stored in the \textit{object memory}. Specifically, object memory constitutes a feature table that connects object visual features with unique object identifiers, and a SQL database that stores the object(person) occurrence information across the video. Details on the construction can be found below: 

\definecolor{light_green}{RGB}{156,220,183}
\definecolor{pink}{RGB}{221,122,208}

\begin{figure}[t]
    \centering
    \includegraphics[width=0.9\textwidth]{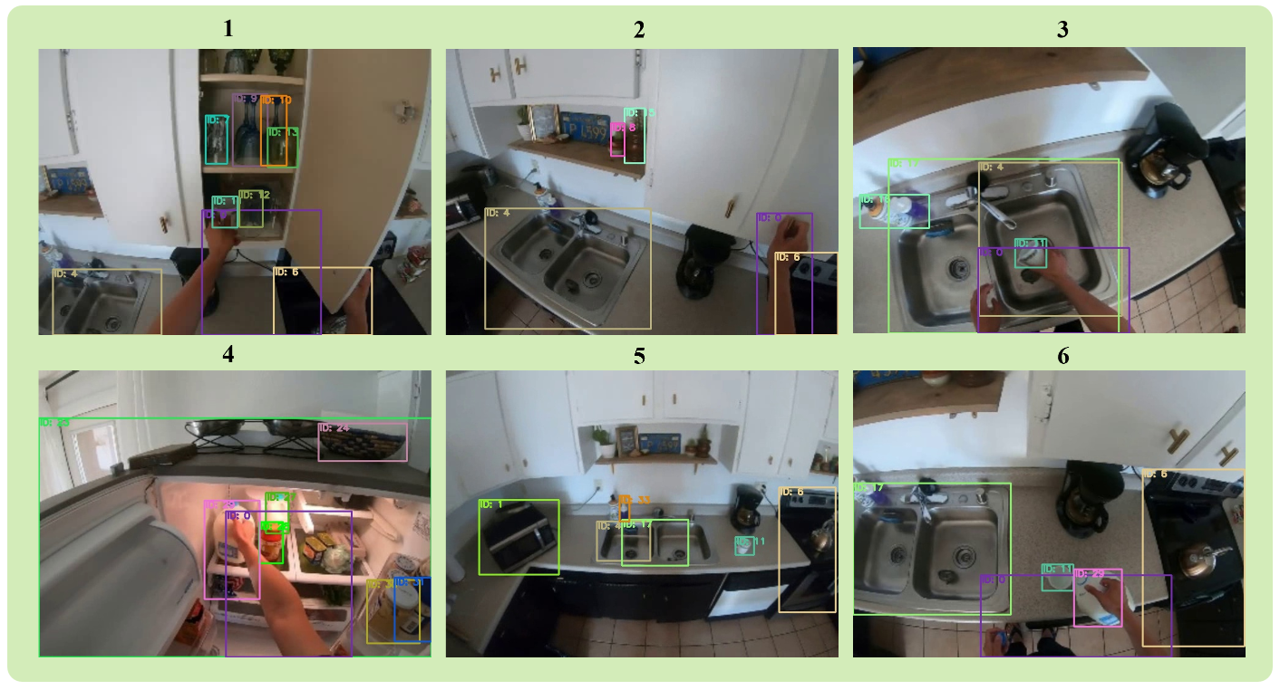}
    \caption{A visualization of object tracking and re-ID. 6 frames from a video are displayed in order. The cup (\textcolor{light_green}{light green} box) and the milk bottle (\textcolor{pink}{pink} box) are successfully re-identified in different postures.}
    \label{fig:reid}
\end{figure}

\noindent\textbf{Tracking and re-identification.} At the heart of our object memory construction pipeline is tracking all the objects across the video, and re-identifying (re-ID) previously appeared objects to eliminate object duplication. We pipeline an object detection model RT-DETR~\cite{rtdetr} with a multi-object tracker ByteTrack~\cite{bytetrack} for the object discovery and tracking part. This combination produces tracking IDs, categories, and bounding boxes of the tracked object occurrences in the video frames. In this phase, an object may have multiple tracking IDs due to its multiple occurrences in the video. For the re-ID part, the key idea is to first compute the features of all the object occurrences that have been discovered and tracked, then group them into object IDs based on their feature similarities. 
More specifically, the feature of an object occurrence (a tracking ID) is generated on object images cropped from $10$ randomly sampled frames of the tracking ID; we also follow a recent study~\cite{tong2024eyes} to use an ensemble of CLIP~\cite{radford2021learning} and DINOv2~\cite{oquab2023dinov2} feature similarity to group tracking IDs into object IDs:
\begin{align*}
    \text{CLIP}(i, j) &= \frac{1}{1+\text{exp}[-20*(\text{cosine}(e^{\text{CLIP}}_i, e^{\text{CLIP}}_j)-0.925)]}, \\
  \text{DINOv2}(i, j) &= \frac{1}{1+\text{exp}[-4.1*(\text{cosine}(e^{\text{DINOv2}}_i, e^{\text{DINOv2}}_j)-0.5)]}, \\
  \text{sim}(i, j) &= 0.15*\text{CLIP}(i, j)+0.85*\text{DINOv2}(i, j),
\end{align*}
where $\text{cosine}(\cdot, \cdot)$ denotes cosine similarity, $e^{\text{CLIP}}_i, e^{\text{CLIP}}_j$ and $e^{\text{DINOv2}}_i, e^{\text{DINOv2}}_j$ are the CLIP and DINOv2 features of the tracking ID $i$ and $j$, respectively. The hyperparameters above (coefficients and biases) are tuned with a simple grid search on EgoObjects~\cite{egoobjects}. More details about re-ID can be found in \supp.
An example of how our tracking and re-ID pipeline manages to handle the temporally discontinuous object presence in a kitchen can be found in \cref{fig:reid}. 

\noindent\textbf{Feature table.} Assuming we've identified all objects (Object IDs) from the video and their object occurrences (tracking IDs) have been confirmed as well. We compute the CLIP feature $f^{s_\text{id}}_{\text{object}}$ of object ID $s_\text{id}$ by averaging the CLIP features of its tracking IDs, and store both the CLIP feature and the object ID in a table. This allows us to use free-form language queries (\eg ``red cup'') to search for objects in the video.

\noindent\textbf{SQL database.} Further, we build a relational database with three fields: object ID $s_\text{id}$, object category $s_\text{category}$, and indices of video segments $\{I_1, \dots, I_t\}$ where the object has appeared. Later, this database can be queried using SQL code and support sophisticated querying logic.

\begin{algorithm}[t] 
    \caption{\label{alg:memory}Memory construction of \method.}  
    \KwIn {video $V$, video captioning model \texttt{video\_cap($\cdot$)}, video embedding model \texttt{video\_emb($\cdot$)}, text embedding model \texttt{text\_emb($\cdot$)}, video object tracker with re-identification \texttt{object\_track\_reid($\cdot$)}} 
    \KwOut {temporal memory $\mathcal{M}_T$, object memory $\mathcal{M}_O$}
    Initialize $\mathcal{M}_T = \varnothing$, $\mathcal{M}_O = \varnothing$\;
    Slicing video into $n$ short segments $V = [v_1, v_2, ...v_n]$ (each segment spans approximately 2 seconds)\;
    \For {$v_i$ in $[v_1, v_2, ...v_n]$}{
        $s_\text{caption}$ $\leftarrow$ \texttt{video\_cap($v_i$)}\;
        $e_\text{video}$ $\leftarrow$ \texttt{video\_emb($v_i$)}\; 
        $e_\text{text}$ $\leftarrow$ \texttt{text\_emb($s_\textnormal{caption}$)}\;
        $\mathcal{M}_T = \mathcal{M}_T + (s_\text{caption}, e_\text{video}, e_\text{text})$
    }
    results $\leftarrow$ \texttt{object\_track\_reid($V$)}\;
    \For {$S$ in $\textnormal{results}$}{
        $s_\text{id}, s_\text{category}, \{I_1, \cdots I_k\}, f^{s_\text{id}}_{\text{object}}$ $\leftarrow$ $S$~~~//See \cref{sec:method_object_mem}\;
        $\mathcal{M}_O = \mathcal{M}_O + (s_\text{id}, s_\text{category}, \{I_1, \cdots I_k\}, f^{s_\text{id}}_{\text{object}})$\;
    }
    return $\mathcal{M}_T$, $\mathcal{M}_O$\;
\end{algorithm}  

\begin{algorithm}[t] 
    \caption{\label{alg:inference}Inference of \method.}  
    \KwIn {task instruction $q$, temporal memory $\mathcal{M}_T$, object memory $\mathcal{M}_O$, LLM \texttt{LLM($\cdot$)}, a set of tools (see \cref{sec:method_tool})}
    \KwOut {response $a$}
    Initialize history $h=[q]$\;
    Initialize inference step count $c = 0$\;
    \While {$c$ < MAX\_STEP}{
        action, input $=$ \texttt{LLM($h$)}\;
        \If {action == "caption\_retrieval"}{
            $t_\text{start}$, $t_\text{end}$ $\leftarrow$ input\;
            results $\leftarrow$ \toolA\xspace\texttt{caption\_retrieval($t_\textnormal{start}, t_\textnormal{end}, \mathcal{M}_T$)}\;
        }
        \ElseIf {action == "segment\_localization"}{
            $s_\text{query}$ $\leftarrow$ input\;
            results $\leftarrow$ \toolB\xspace\texttt{segment\_localization($s_\textnormal{query}, \mathcal{M}_T$)}\;
        }
        \ElseIf {action == "visual\_question\_answering"}{
            $s_\text{question}$, $t_\text{target}$ $\leftarrow$ input\;
            results $\leftarrow$ \toolC\xspace\texttt{visual\_question\_answering($s_\textnormal{question}, t_\textnormal{target}$)}\;
        }
        \ElseIf {action == "object\_memory\_querying"}{
            $s_\text{query}$ $\leftarrow$ input\;
            results $\leftarrow$ \toolD\xspace\texttt{object\_memory\_querying($s_\textnormal{query}, \mathcal{M}_O$)}\;
        }
        \ElseIf {action == "stop"}{
            break\;
        }
        $h = h + [\text{(action, input, results)}]$\;
        $c = c + 1$\;
    }
    return $a = \text{\texttt{LLM($h$)}}$\;
\end{algorithm}

\subsection{Tools and Inference}\label{sec:method_tool}
Compared to counterparts that offer a large collection of tools and usually result in ambiguity in tool calling and complex tool-use pipeline, in \method, our design principle is to provide a minimal but sufficient tool set with a focus on querying the memory. We find this simplifies the inference procedures as well as leads to better performances. We consider the following tools (\toolA\toolB\toolC\toolD):

\noindent\textbf{\toolA\xspace Caption retrieval.} The goal is to extract the captions from specified video segments. Concretely, given the temporal memory $\mathcal{M}_T$, a start and an end time step $t_\text{start}$ and $t_\text{end}$ as arguments, the tool \texttt{caption\_retrieval($\cdot$)} simply retrieves these captions from the temporal memory directly. Due to the context limit, the longest time window allowed is $15$ segments, \ie $t_\text{end} < t_\text{start}+15$.

\noindent\textbf{\toolB\xspace Segment localization.} The goal is to localize a video segment given a text query $s_\text{query}$. The tool \texttt{segment\_localization($\cdot$)} will compare the text feature of $s_\text{query}$ against the video features in the temporal memory $\mathcal{M}_T$. Specifically, we consider an ensemble of the query--video similarity (made possible by ViCLIP~\cite{viclip}, a pretrained video-text CLIP model) and the query--caption similarity (both text features are computed by \texttt{text-embedding-3-large} offered by OpenAI). 
Top-5 video segments will be returned by this tool.

\noindent\textbf{\toolC\xspace Visual question answering.} The goal is to answer a given question $s_\text{question}$ about a short video segment at time $t_\text{target}$, allowing to gather extra information that is not covered by the captions in temporal memory or states in object memory. Concretely, we run Video-LLaVA~\cite{lin2023video} when the tool \texttt{visual\_question\_answering($\cdot$)} is called.

\noindent\textbf{\toolD\xspace Object memory querying.} The goal is to perform sophisticated information retrieval about objects that appeared in the video from the object memory $\mathcal{M}_O$. Specifically, when calling the tool \texttt{object\_memory\_querying($\cdot$)} with a text query $s_\text{query}$ (\eg ``How many red cups did I take out from the fridge?''), relevant object descriptions will first be extracted from the query (\eg ``red cup''); next, we compare the text feature of the descriptions (obtained from CLIP~\cite{radford2021learning}) against the object features from the feature table in $\mathcal{M}_O$ to obtain the object IDs likely correspond to the descriptions; finally, the LLM will write SQL code based on both $s_\text{query}$ and the retrieved object IDs to query the database in $\mathcal{M}_O$ and obtain the needed information (segments that the objects appeared, \etc). After being further processed by the LLM, a response to $s_\text{query}$ will be returned.

\begin{figure}[t!]
    \centering
    \includegraphics[width=0.9\textwidth]{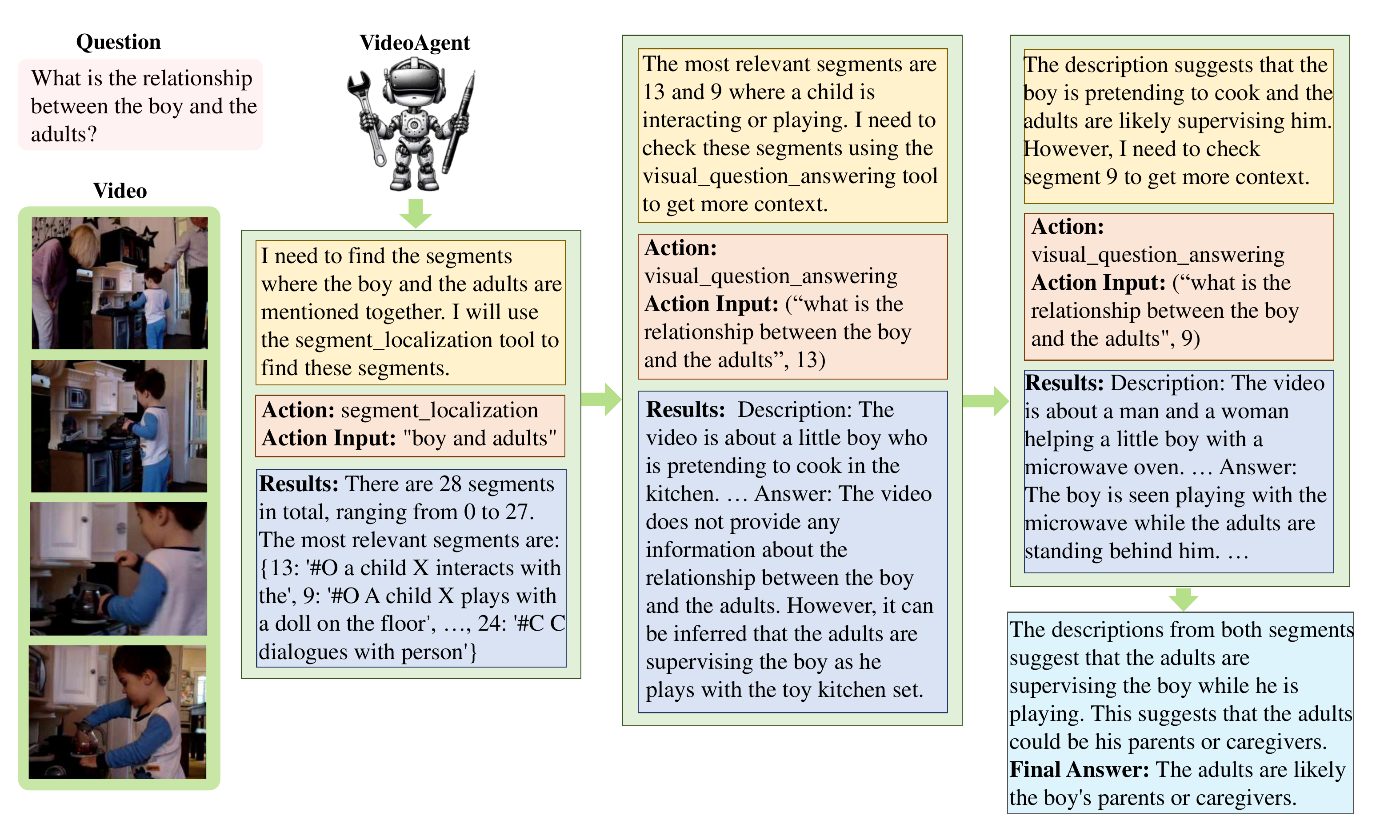}
    \caption{An examples of the \method inference. Given a question, \method executes multiple tool-use steps until it reaches the answer. The yellow, red, and blue blocks in each step denote the chain of thought, action to be taken, and results of tool use. }
    \label{fig:inference_example}
\end{figure}

The inference procedure of \method is rather straightforward. Starting with a history buffer $h$ initialized with the input query $q$, \method decides which tool to use, calls the tool with the produced arguments, appends the results to the buffer, and repeats until it decides to stop or a maximum number of steps is reached. Finally, a response will be made based on the content in the history buffer. We provide an example of this procedure in \cref{fig:inference_example}. \method is implemented using LangChain%
\footnote{\url{https://www.langchain.com/}} with GPT-4 as the main LLM.

\section{Capabilities and Analysis}
\label{exp}
We evaluate \method on various long-form video understanding benchmarks including EgoSchema (\cref{sec:exp_egoschema}), Ego4D Natural Language Queries (\cref{sec:exp_ego4d}), WorldQA (\cref{sec:exp_worldqa}) and NExT-QA (\cref{sec:exp_nextqa}), and  the performances are compared against state-of-the-art end-to-end multimodal LLMs and multimodal agents.

\subsection{EgoSchema}\label{sec:exp_egoschema}
\noindent\textbf{Overview.}
EgoSchema \cite{mangalam2024egoschema} is a benchmark that contains around $5000$ questions about long-form videos. 
The questions typically involve challenging video-level reasoning such as ``describe the general activity in the room and how the different characters and their actions contribute to this environment''.
\method is both tested on the full $5031$-question test set and the official $500$-question subset. The comparative methods include SeViLA~\cite{sevila}, Video-LLaVA~\cite{suris2023vipergpt}, mPLUG-Owl~\cite{ye2023mplug}, ViperGPT~\cite{zhang2023simple}, LLoVi~\cite{llovi}, FrozenBiLM~\cite{frozenbilm}, InternVideo~\cite{wang2022internvideo} and Gimini 1.5 Pro \footnote{\url{https://storage.googleapis.com/deepmind-media/gemini/gemini_v1_5_report.pdf}}.

\begin{table}[t]
\centering
\small
\caption{Accuracy results on the EgoSchema dataset. Top row: results on the full EgoSchema test set; Bottom row: results on the EgoSchema 500 subset.}
\label{egoschema_main}
\setlength{\tabcolsep}{1.1mm}{
\begin{tabular}{cccccc}
\toprule
\multicolumn{6}{c}{\textbf{EgoSchema (full set)}}                                         \\ \hline
FrozenBiLM & InternVideo & mPLUG-Owl & LLoVi & Gemini 1.5 Pro  &\method \\ \hline
26.9       & 32.0        & 30.2       & 50.3    & \textbf{63.2}  & 60.2 \\ \hline
\multicolumn{6}{c}{\textbf{EgoSchema (subset, 500 questions)}}                                                           \\ \hline
SeViLA     & Video-LLaVA & mPLUG-Owl & LLoVi   & ViperGPT  & \method         \\ \hline
25.8       & 36.8        & 33.8         & 51.8   & 15.8       & \textbf{62.8}  \\ \bottomrule
\end{tabular}}
\end{table}

\noindent\textbf{Main results.}
In \cref{egoschema_main}, \method significantly outperforms other state-of-the-art video understanding models such as SeViLA and Video-LLaVA to nearly $30$ percent, achieving an accuracy of $62.8$ on the $500$ questions. Besides, \method achieves $60.2$ on the full test set, closing to the performance of Gemini 1.5 Pro. The strong performance of \method on EgoSchema proves that \method
can solve complex video tasks on long-form videos better than multimodal LLMs and agent counterparts.

\noindent\textbf{Unified memory facilitates stronger reasoning.} The questions in EgoSchema are rather complex in terms of the underlying reasoning about the lengthy videos. Therefore, strong spatial-temporal reasoning is essential. What canonical approaches like multimodal LLMs (Video-LLaVA, \etc) or counterpart multimodal agents (ViperGPT) have in common is the lack of a unified memory as a structured representation for the videos. Without such representation, the reasoning has to be either implicit (as in end-to-end models) or quite limited by the available tools (as in ViperGPT), results in worse performances than ours.

\noindent\textbf{Holistic video understanding with flexible tool-use.} Given a typical question such as "how did c's behavior evolve throughout the video, and what stages of engagement with the tasks can you identify?", it is hard to derive a descriptive text from the question and use it for video grounding, which is a common way for multimodal LLMs (SeViLA, \etc) to select limited key frames for the visual input. However, apart from the \toolB\xspace\texttt{segment\_localization}, \method can also use \toolA\xspace\texttt{caption\_retrieval} to grab the main context of the video and decide which segments are critical, therefore tackling this obstacle.

\subsection{Ego4D Natural Language Queries}\label{sec:exp_ego4d}

\noindent\textbf{Overview.}
The task of Ego4D Natual Language Queries \cite{grauman2022ego4d} is to locate a temporal window (9 seconds on average) in the video (9 minutes on average) that can best answer a query. 
\method is evaluated zero-shot with different variants of the \toolB\xspace\texttt{segment\_localization} tool using 1) ViCLIP visual features only; 2) textual features based on LaViLa captions or Ego4D ground-truth narrations; 3) a combination of both textual features and visual features (LaViLa+ViCLIP and Ego4D+ViCLIP, the ensemble weights can be found in \supp). The methods for comparison include 2D-TAN~\cite{2dtan}, VSLNet~\cite{vslnet}, and GroundNLQ~\cite{groundnlq}, which ranked first in Ego4D NLQ challenge 2023.

\begin{table}[t!]
\centering
\small
\caption{Comparison between supervised baselines and \method with different tool implementation variants on Ego4D NLQ validation set.}
\setlength{\tabcolsep}{3.6mm}{
\begin{tabular}{ccccc}
\toprule
\multicolumn{5}{c}{\textbf{EGO4D NLQ Val.}}                                                                                                      \\ \hline
\multicolumn{1}{c|}{\textbf{Method}} & \textit{\textbf{R1@0.3}} & \textit{\textbf{R1@0.5}} & \textit{\textbf{R5@0.3}} & \textit{\textbf{R5@0.5}} \\ \hline
\multicolumn{5}{c}{\textbf{Supervised}}                                                                                                          \\ \hline
\multicolumn{1}{c|}{2D-TAN}          & 5.04                     & 2.02                     & 12.89                    & 5.88                     \\
\multicolumn{1}{c|}{VSLNet}          & 5.45                     & 3.12                     & 10.74                    & 6.63                     \\
\multicolumn{1}{c|}{GroundNLQ}       & \textbf{27.20}           & \textbf{18.91}           & \textbf{54.42}           & \textbf{39.98}           \\ \hline
\multicolumn{5}{c}{\textbf{Zero-Shot (\method with \toolB\xspace\texttt{segment\_localization} variants)}}                                                                     \\ \hline
\multicolumn{1}{c|}{ViCLIP}          & 8.40                     & 3.97                     & 17.36                    & 8.50                     \\
\multicolumn{1}{c|}{LaViLa}          & 10.07                    & 4.19                     & 22.53                    & 10.58                    \\
\multicolumn{1}{c|}{Ego4D}           & 16.41                    & 6.96                     & 31.96                    & 15.01                    \\
\multicolumn{1}{c|}{LaViLa+ViCLIP}   & 11.13                    & 4.76                     & 25.31                    & 12.08                    \\
\multicolumn{1}{c|}{Ego4D+ViCLIP}    & \textbf{17.39}           & \textbf{7.47}            & \textbf{33.05}           & \textbf{15.73}           \\ \bottomrule
\end{tabular}
}
\label{ego4d_0}
\end{table}
\begin{table}[t!]
\centering
\small
\caption{Comparison between two zero-shot approach: \method and LifeLongMemory~\cite{wang2023lifelongmemory} on Ego4D NLQ. *The performances of LifelongMemory on \textit{\textbf{R1@0.3}} and \textit{\textbf{R5@0.3}}, although not reported, must be less or equal than \textit{\textbf{R@0.3}}.}
\begin{tabular}{c|ccc}
\toprule
\textbf{Method}                 & \textit{\textbf{R1@0.3}} & \textit{\textbf{R5@0.3}} & \textit{\textbf{R@0.3}} \\ \hline
LifeLongMemory(Ego4D) &  *                        & *                        & 15.99                   \\
LifeLongMemory(LaViLa) & *                        & *                        & 9.74                    \\
\method(Ego4D)          & \textbf{16.41}                   & \textbf{31.96}                    & -                       \\
\method(LaViLa)         & 10.07                    & 22.53                    & -                       \\ \bottomrule
\end{tabular}
\label{ego4d_1}
\end{table}

\noindent\textbf{Main results.} \cref{ego4d_0} presents the results on the validation set of Ego4D NLQ. A combination of both textual features and visual features in \method results in better video grounding. Although having a performance gap with the supervised GroundNLQ, \method outperforms 2D-TAN and VSLNet and achieves good performance considering its simple architecture and zero-shot characteristics.

\noindent\textbf{Caption features vs. visual features.} It can be inferred from the comparison among ViCLIP, LaViLa and Ego4D that it is more effective to use the caption--query similarities for video grounding than using video--query similarities. Higher quality captions (LaViLa$\rightarrow$Ego4D) will also lead to better performance.

\noindent\textbf{Similarity-based vs. LLM-based localization.}
\cref{ego4d_1} presents a comparison between \method and LifeLongMemory \cite{wang2023lifelongmemory}. Given a query, LifeLongMemory uses GPT-4 to digest and refine the captions of the video segments, and outputs a list of candidate windows to the query based on the captions selected by the LLM. LifeLongMemory adopts a customized $R@0.3$ metric to calculate the proportion of the predictions where at least one out of all the LLM-generated candidate windows achieves an $IoU$ greater than $0.3$ with the ground-truth window. It can be inferred from \cref{ego4d_1} that given the same caption type (Ego4D or LaViLa), the performance of \method on $R1@0.3$ where only $1$ candidate is allowed for a query, has already surpassed the performance of LifeLongMemory on $R@0.3$. By providing $5$ candidates for a query, the performance of \method will exceed LifeLongMemory by more than two-fold. This indicates that similarity-based segment localization is more effective than the LLM-based segment localization.
\begin{table}[t]
\small
\centering
\caption{Results on WorldQA.}
\begin{tabular}{cccccc}
\toprule
\multicolumn{6}{c}{\textbf{WorldQA}} \\ \hline
\multicolumn{1}{c|}{\textbf{Method}} & Video-LLaMA & Video-ChatGPT & Video-LLaVA & GPT-4V & VideoAgent \\ \hline
\multicolumn{1}{c|}{\textbf{Open-Ended}} & 26.80 & 28.51 & 30.15 & \textbf{35.37} & 32.53 \\
\multicolumn{1}{c|}{\textbf{Multi-Choice}} & 4.81 & 13.25 & 35.25 & 32.83 & \textbf{39.28} \\
\bottomrule
\end{tabular}
\label{worldqa_main}
\end{table}

\begin{table}[t]
\centering
\small
\caption{Results on NExT-QA. We compare baselines on both the original full set as reference and the subset (600 questions) due to the evaluation cost.}
\setlength{\tabcolsep}{3.6mm}{
\begin{tabular}{ccccc}
\toprule
\multicolumn{5}{c}{\textbf{NExT-QA}}                                                                                   \\ \hline
\multicolumn{1}{c|}{\textbf{Method}}   & \textbf{Temporal} & \textbf{Causal} & \textbf{Descriptive} & \textbf{Average} \\ \hline
\multicolumn{5}{c}{\textbf{Val. Set}}                                                                                  \\ \hline
\multicolumn{1}{c|}{InternVideo}       & 43.4              & 48.0            & 65.1                 & 49.1             \\
\multicolumn{1}{c|}{SeViLA(zero-shot)} & \textbf{61.3}     & \textbf{61.5}   & \textbf{75.6}        & 63.6    \\
\multicolumn{1}{c|}{TCR(pre-training)}       & -              & -            & -                 & \textbf{66.1}             \\
\hline
\multicolumn{5}{c}{\textbf{Val. Subset (600)}}                                                                               \\ \hline
\multicolumn{1}{c|}{ViperGPT}          &17.0               &19.0             &26.5                  &20.8              \\
\multicolumn{1}{c|}{mPLUG-Owl}         &36.0                   &  41.0               &52.5                      &43.2                  \\
\multicolumn{1}{c|}{Video-LLaVA}       & 42.0              & 53.5            & 65.0                 & 53.5             \\
\multicolumn{1}{c|}{SeViLA(zero-shot)} & 56.0              & 66.5            & 70.0                 & 64.2             \\
\multicolumn{1}{c|}{\method}            & \textbf{60.0}     & \textbf{76.0}   & \textbf{76.5}        & \textbf{70.8}    \\
\bottomrule
\end{tabular}}
\label{nextqa_main}
\end{table}

\subsection{WorldQA}\label{sec:exp_worldqa}
\noindent\textbf{Overview.} WorldQA~\cite{worldqa} is a challenging video understanding benchmark that focuses on using world knowledge and long-chain reasoning to understand a long-form video (typically a 5-minute movie). 
We compared VideoAgent with Video-LLaMA~\cite{zhang2023video}, Video-ChatGPT~\cite{videochatgpt}, Video-LLaVA~\cite{lin2023video} and GPT-4V~\cite{openai2023gpt4} on both generation-based Open-Ended QA and Multi-Choice QA.

\noindent\textbf{Main results.} 
\cref{worldqa_main} shows that VideoAgent surpasses existing open-source multimodal LLMs by a significant margin on both Open-Ended QA and Multi-Choice QA. This can be mainly contributed to the rich world knowledge and the intrinsic reasoning ability of the LLM agent. Moreover, the better accuracy of VideoAgent compared to that of GPT-4V on Multi-Choice QA demonstrates the effectiveness of the structured memory in understanding long-form videos. On the open-ended QA, GPT-4V achieves better results than VideoAgent, mainly because it has video frames as visual conditions for generating better responses.

\subsection{NExT-QA}\label{sec:exp_nextqa}
\noindent\textbf{Overview.}
NExT-QA \cite{xiao2021next} is a benchmark containing
temporal, causal and descriptive multi-choice questions about videos. The accuracy $acc$ is computed for each type of the questions. 
For the reason of cost, we randomly sampled $200$ questions for each type and obtained a subset of $600$ questions in total to test the performance of \method. Methods directly compared with \method on this subset include ViperGPT \cite{suris2023vipergpt}, mPLUG-Owl \cite{ye2023mplug}, Video-LLaVA \cite{lin2023video} and SeViLA \cite{sevila}. The results of three representative methods InternVideo \cite{wang2022internvideo}, SeViLA \cite{sevila} and TCR \cite{korbar2023text} on the full validation set are also provided.

\noindent\textbf{Main results.}
\cref{nextqa_main} shows the main results on NExT-QA. In all, \method achieves the strongest performances among all comparative methods. Particularly, on the challenging causal questions that require strong temporal understanding and reasoning, \method outperforms SeViLA, one of the state-of-the-art models on NExT-QA, for nearly 10 percent. Besides, the comparison between \method and Video-LLaVA, which is used by the \toolC\xspace\texttt{video\_question\_answering} tool, indicates that our \method allows such multimodal LLM to work better as part of the multimodal tool-use agent than being used alone.

\noindent\textbf{Settings for ablation studies.}
We extract $50$ questions for each question type from the $600$-question subset, resulting in a subset of $150$ questions in total, to evaluate the contributions of different components in \method as ablation studies. \cref{nextqa_ablation} shows the performances of 6 ablations of \method, with each equipped with a unique set of tools among 
\toolC\xspace\textit{visual question answering}, \toolB\xspace\textit{segment localization}, \toolA\xspace\textit{caption retrieval} and \toolD\xspace\textit{object memory querying}, denoted as `VQA', `Grounding', `Captions' and  `Database' in \cref{nextqa_ablation}.

\noindent\textbf{The necessity of caption retrieval.}
The \toolA\xspace\textit{caption retrieval} tool lays the foundation for \method since it provides the basic information about the main context of the video.
With \toolA\xspace\textit{caption retrieval} only, \method of type $6$ achieves an average result of $40.7$ already, which is comparable to the performance $43.2$ of the end-to-end video-language model mPLUG-Owl on the $600$-qeustion subset.

\noindent\textbf{Object memory boosts all question types.} The comparison between type $2$ and $3$ indicates that a reliable object memory can substantially help with temporal and causal questions since it offers crucial temporally consistent object information across video segments, facilitates object-related temporal localization, and enhances the agent's understanding of the video. The performance gap between type $4$ and type $5$ suggests that with the object re-ID algorithm, the performance on descriptive questions (mostly about quantity) will be significantly improved, validating the effectiveness of object re-ID.

\noindent\textbf{VQA and segment localization offer the most bonus.} By comparing type $3$ and $6$, it can be seen that simultaneously adding \toolC\xspace\textit{visual question answering} and \toolB\xspace\textit{segment localization} boost the caption-only \method by 22 percent on the average performance, compared to $15.3$ percent boost by adding the object memory (inferred from type $4$ and $6$). Moreover, by switching from Video-LLaVA to GPT-4V in \toolC\xspace\textit{visual question answering} (type $1$ and $2$), the performance will be raised by 3.4 percent, indicating that accurate visual details identified by the powerful VQA model will aid in better question answering performance.

\begin{table}[t]
\centering
\small
\caption{The effectiveness of different components of \method on NExT-QA subset. {\color{green}\checkmark} and {\color{red}\ding{55}} indicates whether or not the tool is included. "w/ re-ID" uses an object memory constructed with re-ID, while "w/o re-ID" uses an object memory that might include duplicated objects.}
\setlength{\tabcolsep}{0.9mm}{
\begin{tabular}{c|cccc|cccc}
\toprule
\textbf{Type} & \textbf{VQA} & \textbf{Grounding} & \textbf{Captions} & \textbf{Database} & \textbf{Tem.} & \textbf{Cau.} & \textbf{Des.} & \textbf{Avg.} \\ \hline
1             & GPT-4V       & \color{green}\checkmark                  & \color{green}\checkmark                 & w/ re-ID             & 64.0              & 78.0            & 82.0                 & 74.7             \\
2             & Video-LLaVA  & \color{green}\checkmark                  & \color{green}\checkmark                 & w/ re-ID             & 60.0              & 74.0            & 80.0                 & 71.3             \\
3             & Video-LLaVA  & \color{green}\checkmark                  & \color{green}\checkmark                 &\color{red}\ding{55}                 & 46.0              & 64.0            & 78.0                 & 62.7             \\
4             & \color{red}\ding{55}            & \color{red}\ding{55}                  & \color{green}\checkmark                 & w/ re-ID             & 48.0              & 52.0            & 68.0                 & 56.0             \\
5             & \color{red}\ding{55}            & \color{red}\ding{55}                  & \color{green}\checkmark                 & w/o re-ID          & 46.0              & 46.0            & 54.0                 & 48.7             \\
6             & \color{red}\ding{55}            & \color{red}\ding{55}                  & \color{green}\checkmark                 & \color{red}\ding{55}                 & 34.0              & 46.0            & 42.0                 & 40.7             \\ \bottomrule
\end{tabular}}
\label{nextqa_ablation}
\end{table}

\section{Related Work}
\subsection{Multimodal \ac{llm} for video understanding}
Since \ac{llm} have demonstrated an excellent ability to process and understand natural language~\cite{openai2023gpt4,jxma_llm_vla_vlm_mas_multiagent_2023}, several recent works have explored extending them to multimodal setting, especially for images and videos~\cite{shafiullah2022clip,li2023blip,alayrac2022flamingo,lin2023video,team2023gemini,jxma_vlm_multimodal_2023}. %
LaViLa~\cite{lavila} manages to create a massive and diverse set of text as automatic video narrators for video-text contrastive representation pretraining. 
Video-LLaMA~\cite{zhang2023video} enables video comprehension by capturing the temporal changes in visual scenes and integrating audio-visual signals for better cross-modal training. 
As we discussed in \cref{sec:intro}, many of these multimodal foundation models could struggle with long-form video understanding. To remedy this, 
LSTP~\cite{wang2024lstp} utilize spatial and temporal sampler modules to extract optical flow based temporal features and aligned spatial relations from the video to achieve long-form video understanding; Gemini~\cite{team2023gemini} scales the multimodal models to longer videos with tens of thousands of TPUs and massive private video-text datasets. Albeit the prompt progress made by these end-to-end models, prohibitive computation costs and the inherent limitation of the transformer on long-form videos remain significant in applying these end-to-end learned multimodal foundation models to video understanding.

\subsection{Multimodal tool-use agents for video understanding}
Another line of research focuses on augmenting \ac{llm} with a set of \textbf{tools} to solve multimodal tasks without costly training. In particular, \ac{llm} within these \textbf{multimodal agents} are prompted to produce a step-by-step plan to address the original task, and interactively invoke several multimodal foundation models (``tools''), \eg captioning, VQA, \etc. 
VisProg~\cite{gupta2023visual} pilots this direction by equipping the GPT-3 planner with a large collection of visual tools, solving complex real-world visual reasoning problems. 
Applying these agents to video understanding requires careful design as many of the tool models do not guarantee generalization to videos. LifeLongMemory~\cite{wang2023lifelongmemory} employs natural language video narrations to create a text-based episodic memory and prompt \ac{llm} to reason and retrieve required information for the downstream task. 
DoraemonGPT~\cite{yang2024doraemongpt} introduces a sophisticated prompting strategy with Monte Carlo Tree Search (MCTS) to invoke both tools and a structured memory to solve video understanding tasks. These multimodal agents have great potential but so far they mostly struggle with attaining on-par performances to their end-to-end foundation model counterparts on common benchmarks, likely due to the complicated pipelines and lack of video-specific design.

\section{Conclusions}

We've presented \method, a multimodal tool-use agent that reconciles several foundation models with a novel unified memory mechanism for video understanding. Compared to end-to-end multimodal LLMs and tool-use agent counterparts, \method adopts a minimalist tool-use pipeline and does not require expensive training, while offering comparable or better empirical results on challenging long-form video understanding benchmarks including EgoSchema, Ego4D NLQ, WorldQA and NExT-QA. Possible future direction includes more exploration of real-world applications in robotics, manufacturing, and augmented reality.

\section*{Acknowledgements}
We thank the anonymous reviewers for their constructive suggestions. Their insights have greatly improved the quality and clarity of our work. This work was partly supported by the National Science and Technology Major Project (2022ZD0114900).

\bibliographystyle{splncs04}
\bibliography{main}

\newpage
\appendix

In this Appendix, we will first detail the implementation of object re-ID method. Then, the tasks included in \method and the corresponding models will be listed, followed by the experiment settings of \method and all the comparative methods. Finally, cases of the inference of \method will be illustrated.

\section{Object Re-Identification}

Based on the tracking results, object re-identification (re-ID) aims at merging the occurrences of the same object in different period (diverse tracking IDs). The following algorithm shows the procedure of object re-ID. It receives a set of tracking IDs, and output a set of Re-ID groups, where each Re-ID group contains several tracking IDs that belong to the same object, representing a unique object ID in the database. 
\begin{algorithm}[h] 
    \label{algo_reid}
    \caption{Object Re-Identification by Grouping.}  
    \KwIn {video $V$, tracking IDs $\{t_1, t_2, ..., t_n\}$}
    \KwOut {a list of RE-ID groups $G=\{U_1, U_2, ..., U_m\}$}
    Initialize tracking IDs $T = \{t_1, t_2, ..., t_n\}$ to be examined\;
    Initialize the set of re-ID groups $G = \{\}$\;
    
    \For {frame $f$ in $V$}{
        \For {$t_i$ appears in $f$ and $t_i \in T$}{
            \For {Re-ID group $U$ in G}{
                \If {$\forall t_j \in U,\text{share-no-frame}(t_i, t_j)$ and $\forall t_j \in U,\text{sim}(t_i, t_j) > 0.5$ and $\exists t_j \in U, \text{sim}(t_i, t_j) > 0.62$}{remove $t_i$ from $T$\; add $t_i$ to $U$\; break\;}
            }
            \If {$t_i \in T$}{
                remove $t_i$ from $T$\;
                create a new group $U=\{t_i\}$\;
                add $U$ to $G$\;
            }
        }
    }
    output $G$\;
\end{algorithm}  

For each video frame, the algorithm checks every tracking ID in the frame that has not been examined and try to assign it to any existing Re-ID group. A tracking ID $t_i$ should satisfy three conditions in order to be merged to a Re-ID group $U$: 1) It should not co-exist with any tracking IDs in $U$, since the same object only has one bounding box in each frame; 2) It should has $\text{sim}(t_i, t_j)>0.5$ for all $t_j$ in $U$; where $\text{sim}$ refer to the CLIP and DINOv2 feature similarity in the paper; 3) At least one tracking ID $t_j$ in group $U$ satisfies $\text{sim}(o_i, o_j)>0.62$. If the tracking ID $t_i$ cannot be merged to any existing Re-ID group, then the algorithm will spare a new re-ID group initialized with $t_i$. The results of object re-ID are used to construct the SQL database, with each re-ID group corresponding to a unique object ID in the database. 

\section{Tasks and Models}
\cref{tool_models} shows the different tasks in \method and their corresponding models. For each task, the granularity level of the details is also shown. For instance, in the task of segment captioning, the details of captions usually include the actions of the characters and the primary objects in the video that the characters are interacting with.
\begin{table}[h]
\centering
\caption{The methods and the granularity-level of the extracted information in different tasks.}
\label{tool_models}
\small
\begin{tabular}{ccc}
\toprule
\multicolumn{1}{c|}{\textbf{Task}}             & \multicolumn{1}{c|}{\textbf{Method}}               & \textbf{Detail Granularity} \\ \hline
\multicolumn{3}{c}{\textbf{Memory}}                                                                                               \\ \hline
\multicolumn{1}{c|}{Segment Captioning}        & \multicolumn{1}{c|}{LaViLa}                        & action, primary object      \\
\multicolumn{1}{c|}{Object Tracking}           & \multicolumn{1}{c|}{RT-DETR+ByteTrack}             & object category             \\
\multicolumn{1}{c|}{Object Re-ID}              & \multicolumn{1}{c|}{CLIP+DINOv2}                   & object feature              \\ \hline
\multicolumn{3}{c}{\textbf{Tools}}                                                                                                \\ \hline
\multicolumn{1}{c|}{Video Grounding}           & \multicolumn{1}{c|}{ViCLIP+Text-Embedding-3-Large} & action, primary object      \\
\multicolumn{1}{c|}{Visual Question Answering} & \multicolumn{1}{c|}{Video-LLaVA}                   & action, object              \\ \bottomrule
\end{tabular}
\end{table}

\newpage
\section{Experiment Settings}

\subsection{Settings of \method}
\subsubsection{Prompt of \method}
The tool-use capabilities of the LLM (GPT-4) is facilitated using LangChain\footnote{\url{https://www.langchain.com/}}. The LLM is prompted by the following text for the video question answering task.

\begin{tcolorbox}[breakable]
You are tasked with answering a multiple-choice question related to a video. The question has 5 choices, labeled as 0, 1, 2, 3, 4. The video is segmented into 2-second segments, each with an integer ID starting from zero and incrementing in chronological order. Each segment has a caption depicting the event. 
There is an object memory that records the appearing objects in each segment. The object memory is maintained by another agent.
You have access to the following tools:
~\\

\{\textbf{tools}\}
~\\

ATTENTION: 

1. the segment captions with prefix `\#C' refer to the camera wearer, while captions with prefix `\#O' refer to someone other than the camera wearer.

2. You can use both `visual\_question\_answering' and `object\_memory\_querying' to answer questions related to objects or people.

3. The `visual\_question\_answering' may have hallucination. You should pay more attention to the description rather than the answer in `visual\_question\_answering'.

4. The input to the tools should not contain the name of any other tool as well as the token '.

5. Its easier to answer the multiple-choice question by validating the choices.

6. If the information is too vague to provide an accurate answer, make your best guess.
~\\

Use the following format:
~\\

Question: the input question you must answer

Thought: you should always think about what to do

Action: the action to take, should be one of [\{tool\_names\}]

Action Input: the input to the action

Observation: the result of the action... (this Thought/Action/Action Input/Observation can repeat N times)

Thought: I now know the final answer

Final Answer: the correct choice label (0, 1, 2, 3, 4) to the original input question
~\\

Begin!
~\\

Question: \{\textbf{input}\}
Thought: \{\textbf{agent\_scratchpad}\}

\end{tcolorbox}

In the above prompt format, \textbf{tools} refer to a set of tool names and their functional description, including:
\begin{center}
    \fcolorbox{black}{gray!10}{\parbox{0.95\linewidth}{
\texttt{caption\_retrieval}: Given an input tuple (start\_segment, end\_segment), get all the captions between the two segment IDs, 15 captions at most. end\_segment<start\_segment+15.
~\\

\texttt{segment\_localization}: Given a single string description, this tool returns the total number of segments and the top-5 candidate segments with the highest caption-description similarities.
~\\

\texttt{visual\_question\_answering}: Given an input tuple (question, segment\_id), this tool will focus on the video segment starting from segment\_id-1 to segment\_id+1. It returns the description of the video segment and the answer of the question based on the segment.
~\\

\texttt{object\_memory\_querying}: Given an object-related question such as `what objects are in the video?', `how many people are there in the video?', this tool will give the answer based on the object memory. This tool is not totally accurate.
}}
\end{center}
\textbf{input} refers to the multiple-choice question input, including a question and $5$ options. \textbf{agent\_scratchpad} is a list maintained by LangChain that stores the intermediate steps of the agent.

\subsubsection{Object Memory Querying}
\label{memory}
The \texttt{object\_memory\_querying} tool is achieved by another LLM agent (GPT-4) specialized in SQL writing, equipped with the following tools:
\begin{itemize}
    \item database\_querying($program$): return the results from the object memory database by executing the SQL $program$.
    \item open\_vocabulary\_retrieval($description$): return the possible object IDs that satisfy the object $description$.
\end{itemize}
Given an object-related query raised by the central agent, the memory agent will get the relevant object IDs by open vocabulary retrieval, translate the query into SQL program, fetch the results from the database in object memory by running the SQL program, and return the natural language answer to the central agent.

\subsubsection{Experiment Settings of \method}
For NExT-QA and EgoSchema, we use the above prompt for testing the performance of \method. For Ego4D NLQ, the ensemble proportion of video-text and text-text similarities for LaViLa+VICLIP is 18:11, and that for Ego4D+ViCLIP is 7:8. The ensemble proportions is found by grid search on the training set of Ego4D NlQ according to the maximal overall performance on \textit{R1@0.3}, \textit{R1@0.5}, \textit{R5@0.3} and \textit{R5@0.5}.

\subsection{Settings of Comparative Methods}
In the experiments, we test the performance of the following methods by our own. 
The experiment settings for different comparative methods are detailed as follows.
\begin{itemize}
\item SeViLA: The default settings provided in their code are adopted for evaluation. The video frame number is set to $32$, and the key frame number is set to $4$.
\item Video-LLaVA: The default settings provided in their code are adopted for evaluation. The input frame number is set to $8$.
\item mPLUG-Owl:  We follow the evaluation procedure in EgoSchema dataset paper \cite{mangalam2024egoschema},  which prompts mPLUG-Owl by `Given question <question text>, is
answer <answer text> correct?’ along with the video frames. The option with the
highest softmax score of the token ‘Yes’ in the output text will be viewed as the answer of mPLUG-Owl. The input frame number is set to $5$ according to the best mPLUG-Owl settings provided in EgoSchema dataset paper.

\item ViperGPT: GPT-3.5 is adopted as the code generator. $4$ frames are uniformly sampled from the video and the generated code is run on the $4$ frames to gather information for answering the question.
\end{itemize}

\section{Case Study}

In this section, the successful cases of \method on both NExT-QA and EgoSchema are illustrated. In each step towards the final answer, the LLM first reasons about the action to take, and then outputs the action (tool) and its input. The tool will be executed and return the textual results to the LLM. This procedure will iterate until the LLM gets the final answer.

\newpage
\subsection{Case 1}
In this case, the LLM uses the tool \texttt{segment\_localization}, \texttt{caption\_retrieval} and \texttt{visual\_question\_answering} to answer the question. Due to the man in red only occupies a tiny area of the screen, Video-LLaVA does not find the man and produce an answer with hallucination to the question "what does the man do next". However, the LLM will synthesize all the information and produce a correct answer. The video can be found here\footnote{\url{https://youtu.be/5tCWCmCWJKw?si=-IKmlA20_2SqaI_W}}.
\begin{figure*}[h]
    \centering
    \includegraphics[width=0.7\textwidth]{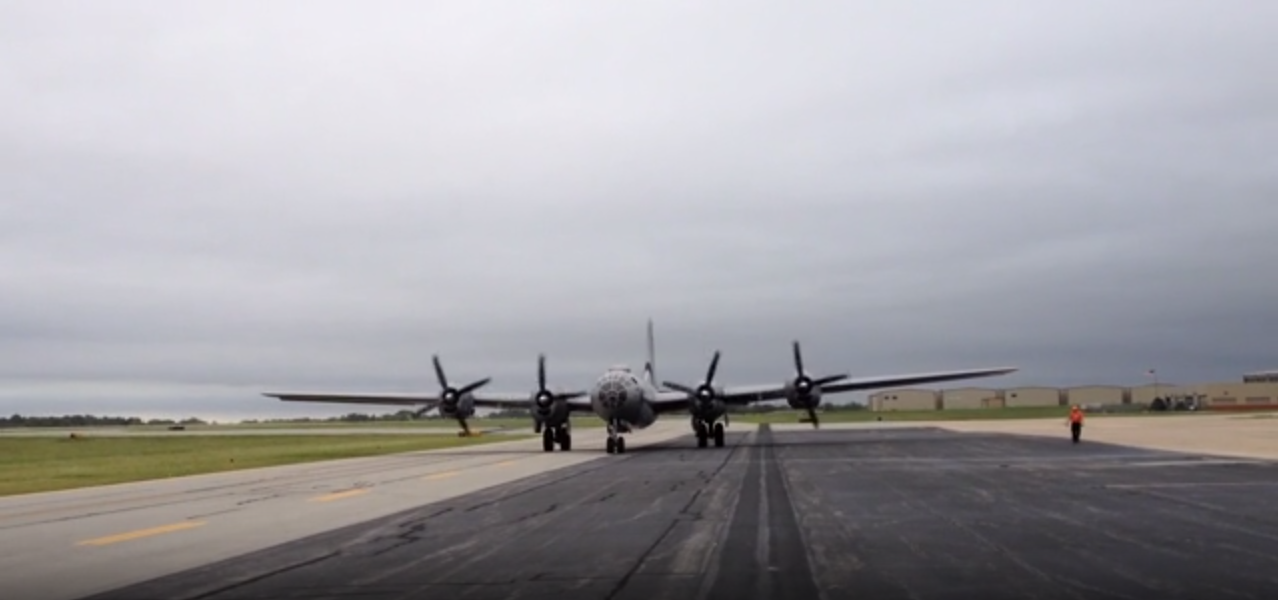}
    \label{plane}
\end{figure*}
\begin{tcolorbox}[breakable]
\textbf{Question}: "what does the man in red do after looking at a distance from the plane at the start"

0: "shakes his head"

1: "happy"

2: "point forward"

3: "count down"

4: "moves away"

~\\

Entering new AgentExecutor chain...
~\\

I need to find the segment where the man in red is looking at a distance from the plane. Then I can use the visual\_question\_answering tool to find out what he does next.

\textbf{Action}: segment\_localization

\textbf{Action Input}: "man in red looking at a distance from the plane"

\textbf{Results}: There are 44 segments in total, ranging from 0 to 43. {22: '\#C C stares the drone on the ground', 15: '\#C C looks around the area', 41: '\#C C looks around the area', 39: '\#O A man x adjusts a drone on the', 40: '\#C C stares the drone on the'}
~\\

The segment with the highest similarity to the description is segment 22, but it doesn't mention a plane. The next highest is segment 15, but it also doesn't mention a plane. The segments 39 and 40 mention a drone, which could be the "plane" referred to in the question. I'll check the captions around these segments to get more context.

\textbf{Action}: caption\_retrieval

\textbf{Action Input}: (37, 42)

\textbf{Results}: \{37: '\#C C looks around the parking', 38: '\#C C looks around the', 39: '\#O A man x adjusts a drone on the', 40: '\#C C stares the drone on the', 41: '\#C C looks around the area', 42: '\#C C looks around the'\}
~\\

The captions from segments 37 to 42 show a man adjusting a drone and then staring at it. This could be the "man in red looking at a distance from the plane" referred to in the question. I'll use the visual\_question\_answering tool to find out what he does next.

\textbf{Action}: visual\_question\_answering

\textbf{Action Input}: ("what does the man do next?", 40)

\textbf{Results}: Description: The video is about a large airplane that is parked on the runway. It shows the plane from different angles, including a close-up of the front of the plane.
Answer: The man in the video walks towards the plane and opens the door.
~\\

The man in the video walks towards the plane and opens the door. This action is not listed in the choices. The closest option is "moves away", which could be interpreted as the man moving away from his original position towards the plane.

\textbf{Final Answer}: 4
~\\

Finished chain.
\end{tcolorbox}

\newpage
\subsection{Case 2}
In this case, \texttt{segment\_localization} and \texttt{visual\_question\_answering} are used to answer the questions. Although some details provided by Video-LLaVA are incorrect, the LLM is able to choose the correct answer based on the captions of the video and the results from Video-LLaVA. The video can be found here\footnote{\url{https://youtu.be/vzlPCFqdtQQ?si=X9vATb1ClBVM8oMM}}.

\begin{figure*}[h]
    \centering
    \includegraphics[width=0.7\textwidth]{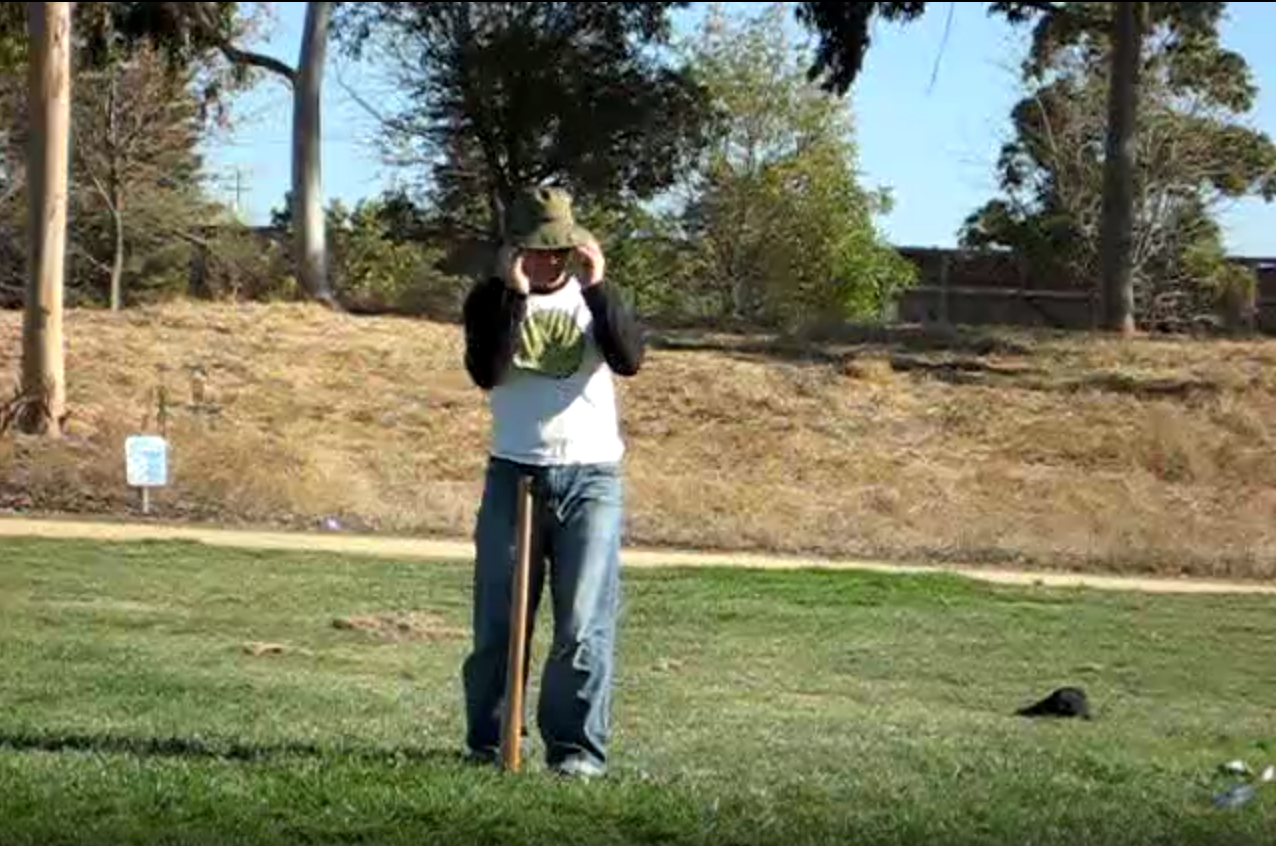}
    \label{bat}
\end{figure*}
\begin{tcolorbox}[breakable]
\textbf{Question}: "how does the boy keep the bat stable on the ground"

0: "lean against his body"

1: "sit on cushion"

2: "put on stone"

3: "swing chair is protected"

4: "wear shoes"
~\\

Entering new AgentExecutor chain...
~\\

The question is asking about a specific action performed by a boy in the video. I need to find the segment where the boy is interacting with the bat and then use the visual\_question\_answering tool to understand how he keeps the bat stable. First, I will use the segment\_localization tool to find the segments where the boy and the bat are mentioned together.

\textbf{Action}: segment\_localization

\textbf{Action Input}: "boy bat"

\textbf{Results}: There are 12 segments in total, ranging from 0 to 11. \{7: '\#O A man Y holds the baseball bat with both', 6: '\#O The man B hits the ball with the rake in his right hand', 9: '\#O man Y holds a head warmer', 10: '\#O A man Z wears a helmet on his head with both hands', 2: '\#O A man X walks around the field'\}
~\\

The segment\_localization tool has returned several segments, but only segment 7 seems to be relevant to the question as it mentions a man (possibly the boy) holding a baseball bat. I will use the visual\_question\_answering tool on this segment to get more details about how the boy keeps the bat stable.

\textbf{Action}: visual\_question\_answering

\textbf{Action Input}: ("how does the boy keep the bat stable?", 7)

\textbf{Results}: Description: The video is about a man who is practicing his archery skills in a park. He is seen holding a bow and arrow and shooting at a target.
Answer: The boy keeps the bat stable by holding it firmly in his hand and keeping it steady. He also uses his body movements to keep the bat stable.
~\\

The visual\_question\_answering tool provided an answer that the boy keeps the bat stable by holding it firmly in his hand and using his body movements. This corresponds to the first option in the multiple-choice question.

\textbf{Final Answer}: 0
~\\

Finished chain.
\end{tcolorbox}

\newpage
\subsection{Case 3}
In this case, the \texttt{visual\_question\_answering} tool contributes most to the successful question answering, since it identifies the wedding ring on the woman's left hand, although it mistakes the woman's action in the video (delivering a talk to others) as 'playing rock paper scissors with a man'. The video can be found here\footnote{\url{https://youtu.be/EFMdUrOELqw?si=6iXTKJCW0x96a7Ap}}.
\begin{figure*}[h]
    \centering
    \includegraphics[width=0.7\textwidth]{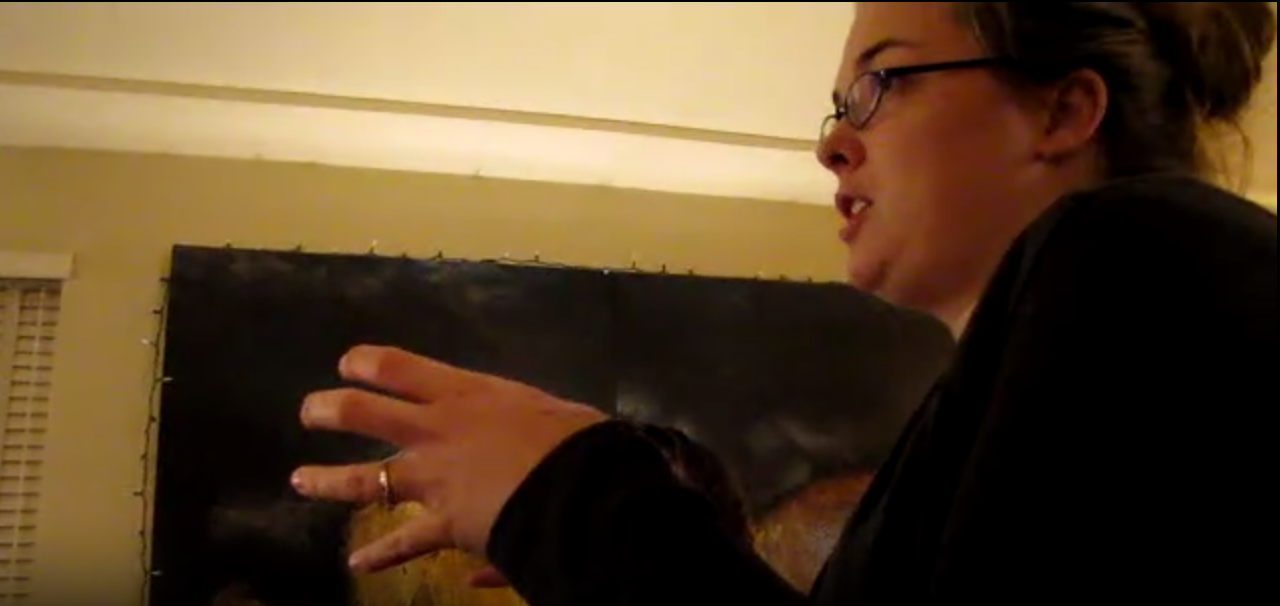}
    \label{ring}
\end{figure*}
\begin{tcolorbox}[breakable]
\textbf{Question}: "how do you know that the woman with black spectacles at the start is married"

0: "wearing a wedding gown"

1: "holding flowers"

2: "wearing a wedding ring"

3: "grandparent and grandchild"

4: "walking with the groom"
~\\

Entering new AgentExecutor chain...
~\\

I need to find the segment where the woman with black spectacles appears at the start of the video. Then I can use the visual\_question\_answering tool to determine if she is married based on the given choices.

\textbf{Action}: segment\_localization

\textbf{Action Input}: "woman with black spectacles at the start"

\textbf{Results}: There are 44 segments in total, ranging from 0 to 43. \{30: '\#O the woman A touches the glasses on her face with her left hand.', 11: '\#O The woman X touches her glasses with her right hand', 14: '\#O person X adjusts her glasses ', 16: '\#O woman X looks at the television', 39: '\#C C stares at a woman'\}
~\\

The woman with black spectacles appears in the first segment (0). Now I will use the visual\_question\_answering tool to determine if she is married based on the given choices.

\textbf{Action}: visual\_question\_answering

\textbf{Action Input}: ("how do you know that the woman is married", 0)

\textbf{Results}: Description: The video shows a man and a woman playing a game of rock paper scissors. The woman wins, and the man gets angry.
Answer: The video does not provide any information about the woman's marital status. However, it is possible to observe that she is wearing a wedding ring on her left hand.
~\\

The woman is wearing a wedding ring which indicates that she is married. This matches 
with choice 2.

\textbf{Final Answer}: 2
~\\

Finished chain.

\end{tcolorbox}

\newpage
\subsection{Case 4}
In this case, the object memory is queried by the LLM (achieved by a memory agent mentioned in \cref{memory}) to answer the quantity of the elephants, and output the correct answer. The video can be found here\footnote{\url{https://youtu.be/OB87EvVi-as?si=DST7FJfGtPmW6ZcB}}.

\begin{figure*}[h]
    \centering
    \includegraphics[width=0.7\textwidth]{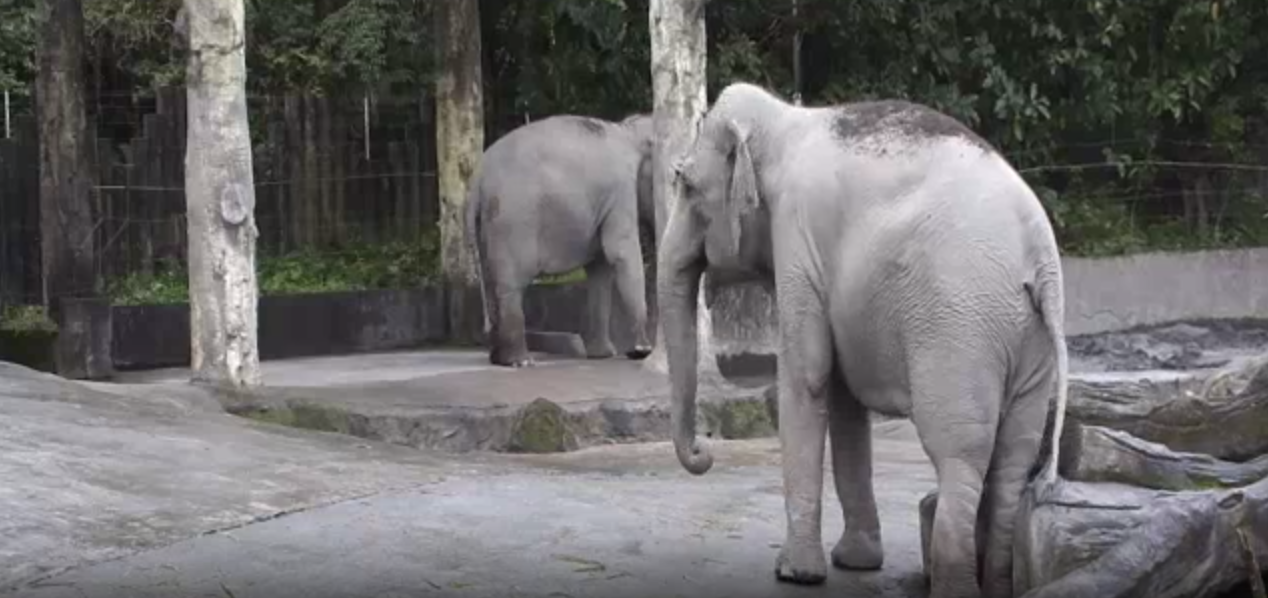}
    \label{elephant}
\end{figure*}
\begin{tcolorbox}[breakable]
\textbf{Question}:"how many elephants are there"

0: "one"

1: "four"

2: "three"

3: "six"

4: "two"
~\\

Entering new AgentExecutor chain..
~\\

I should use the 'object\_memory\_querying' tool to find out how many elephants are in the video.

\textbf{Action}: object\_memory\_querying

\textbf{Action Input}: 'how many elephants are there in the video?'

\textbf{Results}: There are 2 elephants in the video.
~\\

I now know the final answer.

\textbf{Final Answer}: 4
~\\

Finished chain.

\end{tcolorbox}

\newpage
\subsection{Case 5}
This is a typical question in EgoSchema, which requires strong textual understanding and reasoning abilities of the model to correctly answer it. By retrieving the captions, the LLM can gather the information of the ingredients in various segments, and uses \texttt{visual\_question\_answering}` tool on the corresponding segment to investigate the function of each ingredient. The video can be found here\footnote{\url{https://youtu.be/s0LU2K7Ukwo?si=RNEIfZjGCmWeM-1t}}.

\begin{figure*}[h]
    \centering
    \includegraphics[width=0.7\textwidth]{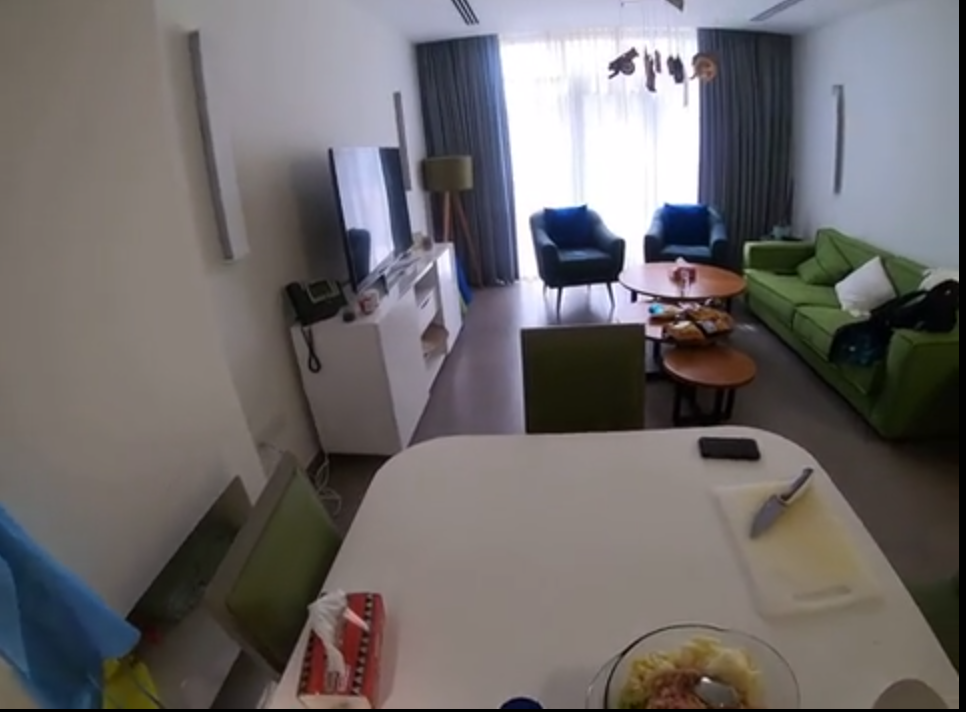}
    \label{dish}
\end{figure*}
\begin{tcolorbox}[breakable]
\textbf{Question}: 'Analyze the choice of ingredients and their importance in the video. how do they contribute to the final dish?', 

0: 'The tissue paper is an important ingredient in the dish. it is used to clean up any spills.' 

1: "The can cover, a crucial component in the dish, serves to maintain the minced beef's freshness effectively by protecting it."

2: 'The wax paper, being a crucial ingredient in the dish, is utilized effectively to securely wrap the can to prevent any potential leakage.'

3: 'The minced beef, spring onions, and mayonnaise are all important ingredients in the dish. the minced beef provides protein, the spring onions provide vitamins and minerals, and the mayonnaise provides flavor.'

4: "The transparent glass bowl serves as an important crucial ingredient in the dish's preparation. it is efficiently used to thoroughly mix the various ingredients together well.
~\\

Entering new AgentExecutor chain...

The question is asking about the importance of ingredients in the final dish. I need to find out what ingredients are used in the video and how they contribute to the final dish. I will start by using the caption\_retrieval tool to get an overview of the events in the video.

\textbf{Action}: caption\_retrieval

\textbf{Action Input}: (0, 14)

\textbf{Results}: \{0: '\#C C scoops sauce from the container with the spoon in her right hand', 1: '\#C C pours the tomato paste from the spoon into the bowl of macaroni and cheese with her right hand', 2: '\#C C scoops sauce from the sauce container with the spoon in her right hand', 3: '\#C C pours the spice into the plate of scrambled eggs with the spoon in her right hand', 4: '\#C C applies the ketchup from the spoon in his right hand on the salad in the bowl', 5: '\#C C scoops sauce from the sauce container with the spoon in her right hand', 6: '\#C C pours the tomato sauce into the bowl on the dining table with the spoon in her right hand', 7: '\#C C  pours milk from the bottle in her left hand into the cup of milk in her right hand', 8: '\#C C puts some sauce from the sauce container on the spoon with her right hand.', 9: '\#C C holds the spoon in the sauce container with her left hand.', 10: '\#C C drops the container on the table with her left hand', 11: '\#C C closes the trash', 12: '\#C C walks towards the dining table', 13: '\#C C drops the tissue paper on the table with her left hand', 14: '\#C C picks a tissue paper from the floor with her left'\}
~\\

The captions indicate that the dish involves sauce, tomato paste, macaroni and cheese, spice, scrambled eggs, ketchup, salad, tomato sauce, milk, and tissue paper. The tissue paper seems to be used for cleaning rather than as an ingredient. I need to find out more about how these ingredients are used. I will use the visual\_question\_answering tool to get more details about the use of these ingredients.

\textbf{Action}: visual\_question\_answering

\textbf{Action Input}: ('What is the role of the sauce in the dish?', 0)

\textbf{Results}: Description: The video is about a person preparing a meal in a kitchen. They are seen cutting up food and mixing it in a bowl.
Answer: The sauce is poured over the noodles and mixed in with the other ingredients, adding flavor and texture to the dish.
~\\

The sauce is used to add flavor and texture to the dish. Now, I will use the visual\_question\_answering tool again to find out more about the role of the other ingredients.

\textbf{Action}: visual\_question\_answering

\textbf{Action Input}: ('What is the role of the spice and scrambled eggs in the dish?', 3)

\textbf{Results}: Description: The video is about a person preparing and eating a meal in a spacious and well-lit room.
Answer: The spice and scrambled eggs are added to the dish to enhance the flavor and texture. The spice adds a burst of flavor and aroma, while the scrambled eggs provide a creamy and tender texture that complements the other ingredients.
~\\

The spice and scrambled eggs are used to enhance the flavor and texture of the dish. Now, I will use the visual\_question\_answering tool again to find out more about the role of the remaining ingredients.

\textbf{Action}: visual\_question\_answering

\textbf{Action Input}: ('What is the role of the tomato sauce and milk in the dish?', 6)

\textbf{Results}: Description: The video is about a person preparing and eating a meal in a spacious and well-lit kitchen. The person is seen cutting up food on a table, mixing it in a bowl, and then eating the food.
Answer: The tomato sauce and milk are poured over the ice cream, adding a sweet and tangy flavor to the dish.
~\\

The tomato sauce and milk are used to add a sweet and tangy flavor to the dish. However, the answer seems to be hallucinating about ice cream, which was not mentioned in the captions. I will use the visual\_question\_answering tool again to find out more about the role of the remaining ingredients.

\textbf{Action}: visual\_question\_answering

\textbf{Action Input}: ('What is the role of the tomato paste and macaroni and cheese in the dish?', 1)

\textbf{Results}: Description: The video is about a person preparing and eating a meal in a spacious and well-lit room.
Answer: The tomato paste and macaroni and cheese are added to the dish to enhance its flavor and texture. The tomato paste adds a rich and savory taste, while the macaroni and cheese provides a creamy and comforting element.
~\\

The tomato paste and macaroni and cheese are used to enhance the flavor and texture of the dish. The tomato paste adds a rich and savory taste, while the macaroni and cheese provides a creamy and comforting element. Now, I will use the visual\_question\_answering tool again to find out more about the role of the tissue paper.

\textbf{Action}: visual\_question\_answering

\textbf{Action Input}: ('What is the role of the tissue paper in the dish?', 13)

\textbf{Results}: Description: The video is about a person cleaning a bathroom.
Answer: The tissue paper is used to clean the dish.
~\\

The tissue paper is used for cleaning, not as an ingredient in the dish. Now, I can compare the information I have gathered with the choices provided in the question.

\textbf{Final Answer}: 3
~\\

Finished chain.

\end{tcolorbox}

\end{document}